# AI Models Exceed Individual Human Accuracy in Predicting Everyday Social Norms

**Authors:** Pontus Strimling[1,2], Simon Karlsson[1], Irina Vartanova[1,3], Kimmo Eriksson[1,4]*

[1]Institute for Futures Studies, Stockholm, Sweden
[2]Institute for Analytical Sociology, Linköping University, Norrköping, Sweden
[3]Department of Women's and Children's Health, Uppsala University
[4]School of Education, Culture and Communication, Mälardalen University, Västerås, Sweden

*Corresponding author: Kimmo Eriksson, Mälardalen University, Box 883, SE-72123 Stockholm, Sweden. E-mail: kimmo.eriksson@mdu.se

**Author Contributions**

PS conceived Study 1. KE conceived Study 2. IV collected the LLM data. IV and SK performed the analyses with input from KE. SK drafted a first version of the introduction. KE wrote the full paper. All authors critically reviewed and approved the final manuscript.

**Funding**

This research was funded by the Knut and Alice Wallenberg Foundation (grant no. 2022.0191, recipient PS) and a Mälardalen University AI and Society Research Fellowship (recipient KE).

# Abstract

A fundamental question in cognitive science concerns how social norms are acquired and represented. While humans typically learn norms through embodied social experience, we investigated whether large language models can achieve sophisticated norm understanding through statistical learning alone. Across two studies, we systematically evaluated multiple AI systems' ability to predict human social appropriateness judgments for 555 everyday scenarios by examining how closely they predicted the average judgment compared to each human participant. In Study 1, GPT-4.5's accuracy in predicting the collective judgment on a continuous scale exceeded that of every human participant (100th percentile). Study 2 replicated this, with Gemini 2.5 Pro outperforming 98.7% of humans, GPT-5 97.8%, and Claude Sonnet 4 96.0%. Despite this predictive power, all models showed systematic, correlated errors. These findings demonstrate that sophisticated models of social cognition can emerge from statistical learning over linguistic data alone, challenging strong versions of theories emphasizing the exclusive necessity of embodied experience for cultural competence. The systematic nature of AI limitations across different architectures indicates potential boundaries of pattern-based social understanding, while the models' ability to outperform nearly all individual humans in this predictive task suggests that language serves as a remarkably rich repository for cultural knowledge transmission.

# Introduction

How do cognitive systems acquire and represent knowledge about social appropriateness? This fundamental question in cognitive science involves a tension between different learning mechanisms. Some approaches emphasize top-down processes like explicit instruction and social feedback, while others focus on bottom-up statistical learning from environmental patterns. This tension becomes particularly intriguing when considering whether sophisticated social understanding can emerge purely from pattern recognition in linguistic data, without the lived experiences that many theories consider essential for cultural competence.

Research on statistical learning demonstrates that humans can extract complex regularities from input without explicit instruction, from basic sequence patterns in infancy to sophisticated linguistic structures (Elman, 1990; Saffran et al., 1996). Distributional approaches suggest that meaning and social knowledge can emerge from co-occurrence patterns in language alone, while Bayesian models show how social and pragmatic understanding can arise through probabilistic inference over communicative patterns (Goodman & Frank, 2016; Harris, 1954; Landauer & Dumais, 1997). These distributional and statistical learning approaches also engage with theories of embodied cognition, which propose that conceptual understanding emerges from sensorimotor experience and bodily interaction with the environment (Barsalou, 2008). Strong embodiment theories would predict that social understanding requires direct experience with social situations, making this question particularly theoretically significant. However, moderate embodiment theories allow for abstract social knowledge to be acquired through linguistic input that describes embodied experiences, which could potentially reconcile statistical learning with embodied cognition frameworks.

Now consider large language models (LLMs) like GPT and Gemini. These systems wield powerful artificial intelligence without embodied experience, relying entirely on statistical learning from text. At the same time, they are rapidly taking on important roles in the social world. This situation raises a profound question: can computational systems achieve human-level understanding of our social world without the lived, physical experiences that many theories consider essential?

A particularly interesting case is whether LLMs can understand everyday social norms. These norms are typically unwritten, context-dependent, and culturally variable. They require understanding not just what behaviors are possible, but what is appropriate given specific social contexts—nuanced judgments that humans make largely intuitively. These characteristics of everyday norms make them especially challenging to understand for systems that rely entirely on statistical learning from text.

**Background on AI and Social Reasoning**

Recent research has begun empirically testing AI social reasoning capabilities, but with important methodological limitations for understanding computational social cognition. Much work examines AI performance on high-stakes moral dilemmas (Awad et al., 2018; Takemoto, 2024) or uses binary classifications of social behaviors (Forbes et al., 2021; Ziems et al., 2023). These approaches miss the continuous, nuanced nature of everyday social appropriateness and typically benchmark AI against predetermined "correct" answers rather than the natural variation in human social cognition. Moreover, this research focuses on moral reasoning rather than the everyday norm understanding that governs most social interaction.

Complementing this work, studies have examined LLMs' understanding of everyday social norms in physical settings. Resources such as SOCIAL-CHEM101 (Forbes et al., 2021), NormBank (Ziems et al., 2023), and EgoNormia (Rezaei et al., 2025) test LLMs' grasp of common social behaviors in contexts like public transport and workplace settings. These evaluations reveal mixed results: LLMs perform well in structured scenarios but struggle with subtle contextual dependencies.

However, existing research has a key limitation from a cognitive science perspective. When comparing AI to human judgment, studies typically benchmark AI against predefined correct answers or aggregate human averages. This approach obscures how AI performance compares to typical individual humans and fails to acknowledge the natural variation in human social cognition. Understanding this variation is crucial for theories of cultural competence and social learning.

**The Present Study**

This paper addresses these limitations through a two-part investigation designed to provide a rigorous benchmark of AI's ability to understand everyday social norms—the unwritten rules about how appropriate a given behavior is in a given context. Consider, for example, how appropriate it is to laugh at a job interview, cry on a bus, or read in church. These judgments involve nuanced social understanding that goes far beyond knowing what behaviors are physically possible. Our overall approach shifts the focus from abstract morality and binary classifications to the continuous spectrum of everyday social appropriateness, and from aggregate averages to a novel individual-level performance comparison.

We leverage a large-scale dataset of human appropriateness judgments for 555 everyday behaviors from 555 U.S. participants. This dataset provides both the cultural consensus to predict and the individual variation to benchmark against.

In Study 1, we provide an initial proof-of-concept by evaluating the ability of GPT-4.5 to predict the cultural consensus about everyday norms and benchmarking its performance against individual human participants. This first study establishes whether a sophisticated model of social norms can be learned from text alone.

GPT-4.5 was a state-of-the-art model in early 2025 that was subsequently retired in summer 2025. In Study 2, we examine the robustness and evolution of LLM capabilities to predict everyday norms. We replicate the entire study with several next-generation models released later in 2025, including GPT-5, Claude Sonnet 4, and Gemini 2.5 Pro. This comparative approach allows us to not only test the generalizability of our initial findings but also to explore the qualitative differences in the "social intelligence" of different leading AI systems.

Across both studies, our central research question remains: if an AI system were evaluated as just another participant in a social cognition study, would its performance fall within the range of typical human variation, or would it demonstrate a capability to model the collective norm that exceeds that of a typical individual? As our results will show, the AI's grasp of our social world is not only highly accurate but, in its ability to reflect the collective consensus, demonstrates a predictive accuracy that exceeds the vast majority of individual humans, a finding with profound implications for cognitive science and AI development.

**Methodological Note on Task Comparison**

The tasks for the AI and human participants are not identical: the AI is prompted to perform a meta-cognitive task of predicting the group average, whereas each human provides their own appropriateness rating. Therefore, our test is a direct assessment of the model's ability to extract a collective social signal from its training data, which is distinct from possessing genuine, human-like social understanding. Nonetheless, a comparison between the AI's predictive accuracy and individual humans' deviation from the mean is justified on the theoretical premise that a judgment of social appropriateness is not a statement of personal preference as much as it is an individual's report on their perception of a shared, collective cultural norm. In this view, each human rating is an estimate of this societal standard. Therefore, while the explicit instructions differ, both the AI and human participants are engaged in a process of accessing and representing a collective consensus, allowing for a meaningful comparison of their accuracy relative to that consensus.

This theoretical framework is supported by extensive research in social cognition demonstrating that individual appropriateness judgments are not purely idiosyncratic preferences but rather reflect individuals' estimates of shared cultural standards (Zou et al., 2009). Social psychological research on pluralistic ignorance shows that people consistently attempt to infer others' attitudes when making social judgments (Miller & McFarland, 1987), suggesting that appropriateness ratings inherently involve a meta-cognitive component similar to the explicit prediction task given to AI. Furthermore, dual-process theories of social cognition propose that social judgments often involve both intuitive responses and deliberative estimation of social consensus (Strack & Deutsch, 2004). From this perspective, the AI's explicit prediction task may actually isolate the deliberative component that is present but confounded in human appropriateness ratings. While the tasks differ in their explicit instructions, both fundamentally require accessing and representing cultural knowledge to generate responses that align with societal expectations.

# Study 1

## Methods and Materials

### Human Normative Data

The human data were sourced from a large-scale, pre-registered study of social norms in the United States (Anonymous, 2023), designed to elicit appropriateness ratings for a comprehensive range of everyday behaviors.

**Stimuli.** The stimuli consisted of 555 unique scenarios created by pairing each of 37 everyday behaviors with each of 15 common situations, replicating and extending Price and Bouffard's (1974) pioneering methodology. This systematic combination allows examination of how the same behavior is perceived across different social contexts.

The 37 behaviors were: Argue, Belch, Blow one's nose, Bring a dog, Chew gum, Cry, Curse, Dance, Do a jigsaw puzzle, Drink water, Eat, Fart, Fiddle with one's phone, Fight, Flirt, Hold hands, Jump, Kiss, Laugh, Listen to music on headphones, Mumble, Play cards, Pray audibly, Read, Run, Shout, Sigh, Sing, Sit on the floor/ground, Spit, Talk, Talk on the phone, Take a selfie, Wave to a friend, Whistle, Work on a laptop, and Write.

The 15 situations were: in a bar, on a bus, in church, in class, on a date, on a downtown sidewalk, in a dormitory lounge, in an elevator, at a family dinner, at a football game, at a job interview, at the movies, in one's own room, in the park, and in a restroom.

**Participants.** The study included 555 participants from the United States, recruited via Prolific in January 2023. Recruitment aimed for balanced representation across gender, age, political ideology, and education level, providing a diverse sample for establishing cultural consensus.

**Appropriateness Rating Scale.** Following Price and Bouffard (1974), participants rated the perceived appropriateness of each situated behavior on a ten-point scale from 0 to 9, anchored at 0 = *The behavior is extremely inappropriate in this situation* and 9 = *The behavior is extremely appropriate in this situation*. This continuous scale captures the nuanced nature of social judgments better than binary classifications.

**Procedure.** To prevent participant fatigue while ensuring reliable estimates, each participant rated 50 randomly selected scenarios (10 behaviors × 5 situations). This design provided approximately 40 ratings per scenario, yielding stable measures of collective social norms (median standard error = 0.32, range = 0.04-0.53).

## AI Normative Data

**Model and Parameters.** We used GPT-4.5 Preview ('gpt-4.5-preview-2025-02-27'), which was available through OpenAI's API during our data collection period (May 5-13, 2025), but has since been retired. To ensure consistent yet naturally variable responses, we used: temperature=0.5 (chosen to balance consistency with natural variation), top_p=1.0, max_tokens=4000.

**Prompting Procedure.** For each scenario, the model received a carefully structured prompt: "From various sources in our everyday lives we have all developed a subjective 'impression' or 'feeling' for the appropriateness of any given behavior in a particular situation. Imagine a number of people from the United States rated the appropriateness of [scenario] on the scale from 0 through 9, where 0 = The behavior is extremely inappropriate in this situation and 9 = The behavior is extremely appropriate in this situation. Note that this scenario combines a behavior and a situation (not a behavior and an object). Interpret it as someone performing the behavior during the situation, in the way such a phrase would typically be understood. For example, 'writing on a bus' means writing while riding the bus, not on its surface. Your task is to estimate the average rating that these U.S. respondents would give. Please use numbers with up to two decimals to provide as detailed estimates as possible. Do not write any comments or justifications for the estimates. Write only the estimate."

This prompt was designed to present the exact task given to human participants, ensuring comparable cognitive demands and interpretation. The culturally-specific instruction was essential for testing whether the AI could access U.S. cultural knowledge rather than relying on generic responses.

**Multiple Queries.** To account for the stochastic nature of language models and ensure robust estimates, we queried the model five times for each scenario. This approach balances computational efficiency with statistical reliability, providing data to capture the model's central tendency while accounting for response variability.

# Results

Our analysis examined how accurately GPT-4.5 could predict human judgments of social appropriateness, addressing both aggregate correlation and individual-level performance comparison.

**Correlation with Human Consensus**

The model's continuous-scale predictions showed exceptionally strong correlation with average human ratings, explaining 89% of the variance in human social norms ($R^2 = 0.89$). Figure 1 reveals tight linear correspondence across the full spectrum of appropriateness judgments. To ensure this finding wasn't an artifact of the continuous scale format, we performed a robustness analysis by rounding AI predictions to the nearest integer before re-analysis. The correlation remained virtually unchanged ($R^2 = 0.88$), confirming that the model's accuracy is robust and not dependent on rating scale precision

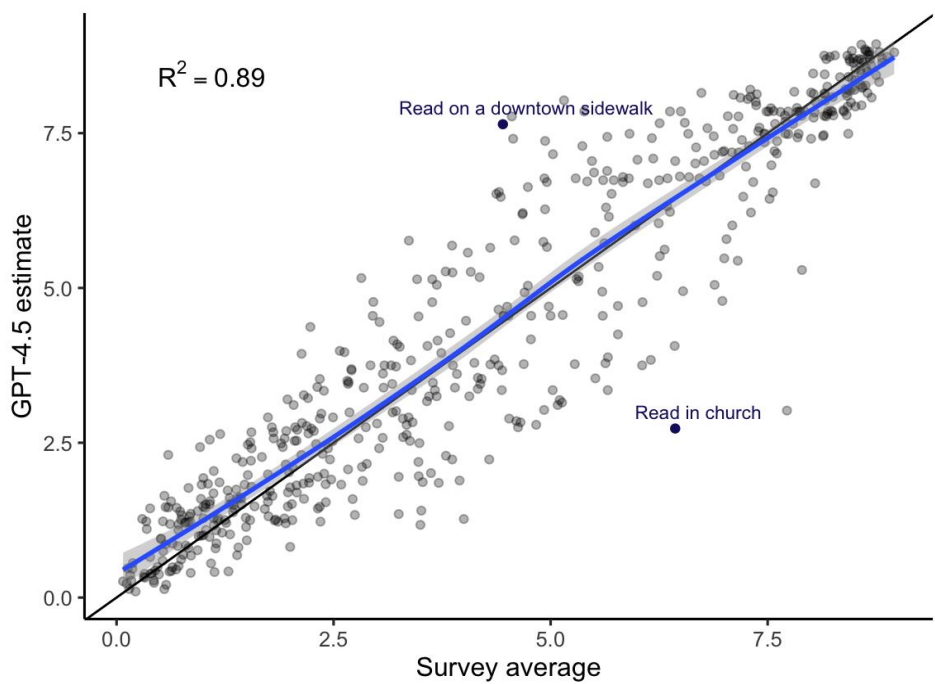

**Figure 1.** Correlation between AI-generated appropriateness ratings and average human ratings for 555 scenarios. Each dot represents a scenario as exemplified by the labeled dots.

**Individual-Level Performance Comparison**

More striking than the aggregate correlation was the AI's performance relative to individual humans. Using Mean Absolute Error (MAE) from the group average as our metric, GPT-4.5's continuous predictions placed it in the 100th percentile of human accuracy, outperforming all 555 participants (Figure 2). Even when constrained to integer ratings matching the human scale, the AI outperformed 553 of 555 human participants (99.6th percentile), demonstrating that its superior performance reflects a genuine and powerful capability for modeling social norms from linguistic data, rather than being an artifact of the rating scale. To ensure these findings were not confounded by differences in scenario difficulty across the subsets rated by different participants, we conducted a complementary analysis comparing each individual's

MAE to the AI's MAE on the identical subset of scenarios and obtained virtually identical results (Supplementary Figure 1).

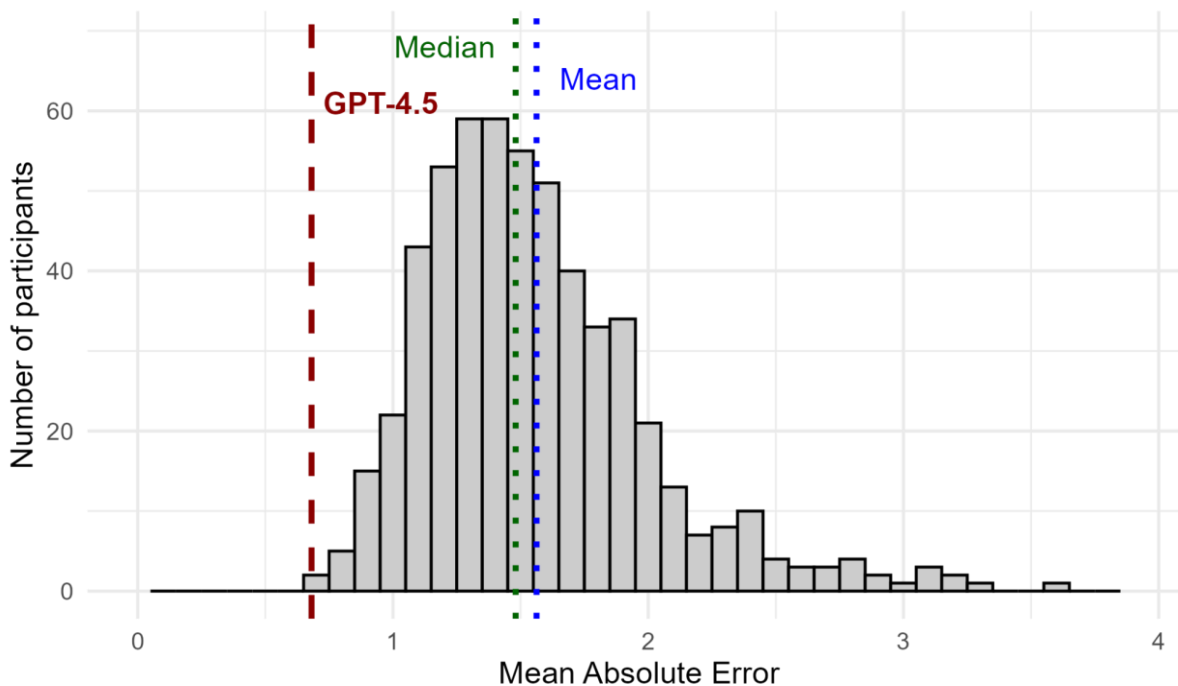

**Figure 2.** Distribution of Mean Absolute Error (MAE) for 555 human participants, with the MAE of GPT-4.5 indicated.

**Analysis of AI-Human Divergences**

While overall correlation was strong, examining specific scenarios where AI judgments diverged from human consensus revealed interesting patterns. Table 1 shows the ten scenarios with largest AI-human differences. The AI systematically underestimated the appropriateness of some context-specific behaviors (e.g., waving to a friend in one's own room, or reading in church) while overestimating the appropriateness of others (e.g., running at a football game, or reading on a downtown sidewalk). These exceptions potentially reflect differences between statistical pattern recognition and experiential knowledge of social contexts.

Table 1: Top 10 Scenarios with Largest AI-Human Divergence.

| Scenario | AI Rating | Human Mean Rating | Difference | Direction |
|---|---|---|---|---|
| Wave to a friend in one's own room | 3.02 | 7.72 | -4.70 | Underestimated |
| Read in church | 2.73 | 6.43 | -3.70 | Underestimated |
| Run at a football game | 7.77 | 4.55 | 3.22 | Overestimated |
| Read on a downtown sidewalk | 7.64 | 4.45 | 3.19 | Overestimated |
| Run in one's own room | 8.03 | 5.15 | 2.88 | Overestimated |
| Kiss at the movies | 7.41 | 4.57 | 2.84 | Overestimated |
| Mumble at the movies | 1.27 | 4.00 | -2.73 | Underestimated |
| Read in class | 5.29 | 7.89 | -2.60 | Underestimated |
| Drink water at a job interview | 7.85 | 5.39 | 2.46 | Overestimated |
| Write at a job interview | 7.37 | 4.94 | 2.44 | Overestimated |

**Discussion of Study 1**

The results of Study 1 provide a powerful proof-of-concept, demonstrating that a sophisticated model of human social norms can be inferred from statistical learning over text alone. The finding that GPT-4.5's accuracy in predicting the U.S. cultural consensus exceeded that of nearly all human participants is a significant finding in itself, supporting bottom-up theories of social cognition. It establishes that the capability for high-fidelity social norm prediction exists in artificial systems.

However, these initial findings raise several crucial questions. Is this remarkable performance an idiosyncratic feature of this specific model, or does it represent a more general capability of state-of-the-art AI? Has this capability evolved or improved in the newer, more powerful models released since our initial data collection in early 2025? Finally, do different AI systems, with their unique architectures and training, exhibit qualitatively different patterns in their social judgments, even if their overall accuracy is high? To address these questions of robustness, evolution, and generality, we conducted a second study.

# Study 2

## Methods and Materials

The methodology for Study 2 was designed to be a direct replication and extension of Study 1. We used the identical human normative dataset and the exact same experimental prompt as detailed in Study 1 to query three new large language models. We selected these models to test generalizability across the most prominent state-of-the-art LLM approaches: OpenAI's latest GPT model, Anthropic's latest Claude Sonnet model, and Google's latest Gemini model. All data for Study 2 was collected between August 10-15, 2025. As in the first study, each of the 555 scenarios was queried five times for each model, and the responses were averaged to obtain a final estimate.

### Models and Parameters

The specific models and their parameters are detailed below.

**GPT-5:** The model version was gpt-5-2025-08-07. We used parameter values top_p=1.0, max tokens = 2048. The GPT-5 API did not support setting a temperature parameter.

**Claude Sonnet 4:** The model version was claude-sonnet-4-20250514. We used parameter values temperature=0.25, max tokens = 2048. We adjusted the temperature value to represent the smaller possible range for the parameter, which is [0, 1] for Claude Sonnet 4 and [0, 2] for GPT-4.5 and Gemini 2.5 Pro.

**Gemini 2.5 Pro**: The model version specified was 'gemini-2.5-pro', which corresponded to the latest stable release during the data collection period. We used parameter values temperature=0.5, max tokens = 4096.

# Results

The analysis for Study 2 aimed to replicate the findings from Study 1 with a new suite of models and to conduct a comparative analysis of their performance and response patterns.

**Correlation with Human Consensus**

All three next-generation models showed a strong positive correlation with the human norm consensus, confirming the primary finding from Study 1. See Figure 3 for an illustration of the results. Note that although all models performed strongly, some were even better than others: $R^2$ values ranged from 0.82 for Claude Sonnet 4 to 0.91 for GPT-5. Furthermore, note that the models exhibited qualitatively different response patterns. While both GPT models and Gemini 2.5 produced fine-grained continuous estimates, Claude Sonnet 4's responses showed a distinct "quantization" effect, clustering heavily at certain values (1.25, 2.35, 3.25, 4.25, 6.75, 7.85, 8.75). This suggests a strong internal bias towards certain numbers that may limit its ability to capture the subtle, continuous nature of human social consensus.

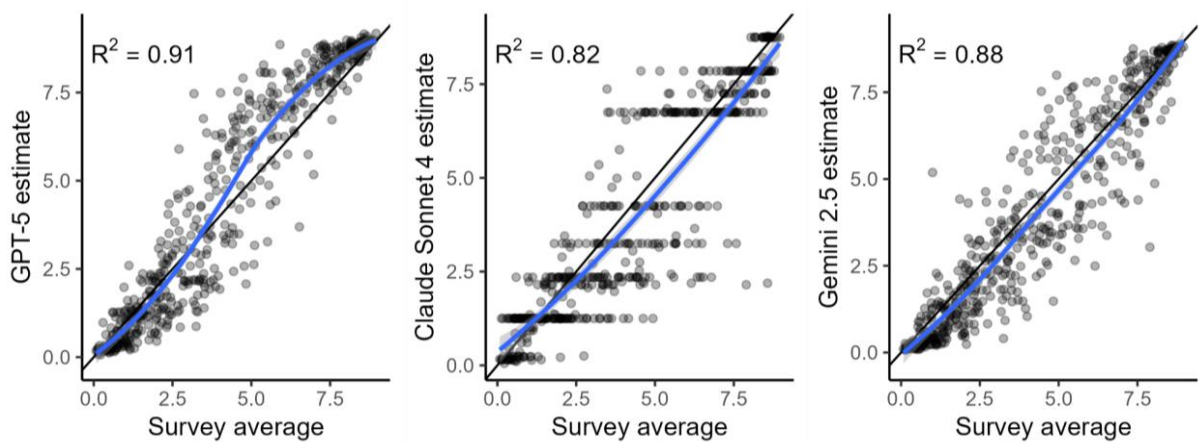

**Figure 3.** Correlation between AI-generated appropriateness ratings and average human ratings for three next-generation models. While all models show strong correlations, GPT-5 demonstrates the highest accuracy, and Claude Sonnet 4 exhibits a distinct 'quantized'

response pattern, with responses clustering at specific numerical values, suggesting different internal representation strategies.

**Individual-Level Performance Comparison**

To compare the models' performance at a finer grain, we benchmarked each model against the 555 individual human participants using Mean Absolute Error (MAE). As shown in Table 2, all models performed at a level exceeding the vast majority of human participants (96% or more). Interestingly, when performance was measured by MAE, GPT-5 did not reach the level of the older model GPT-4.5. As in Study 1, we verified these results were not confounded by differences in scenario difficulty across participant subsets by comparing each individual's MAE to each AI model's MAE on identical scenarios, yielding consistent patterns (Supplementary Figure 1).

**Table 2. Individual-Level Performance Comparison Across All Models.**

| AI model | Continuous estimates | | Integer estimates | |
|---|---|---|---|---|
| | MAE | % of humans outperformed | MAE | % of humans outperformed |
| GPT-4.5 | 0.68 | 100.0 | 0.75 | 99.6 |
| GPT-5 | 0.81 | 99.1 | 0.88 | 97.8 |
| Claude Sonnet 4 | 0.87 | 98.0 | 0.95 | 96.0 |
| Gemini 2.5 Pro | 0.76 | 99.5 | 0.85 | 98.7 |

Note: The results for GPT-4.5 are from Study 1.

The performance differences across models likely reflect distinct architectural and training approaches. Claude Sonnet 4's quantized response pattern suggests it may use more discrete internal representations or have been trained with different optimization objectives that favor certain numerical values. The divergence between GPT-5's at the same time superior correlation but inferior calibration compared to GPT-4.5 indicates that newer models may capture more nuanced relationships between scenarios while sacrificing precise numerical calibration. This dissociation suggests that correlation and calibration may rely on different computational mechanisms, with correlation reflecting the model's understanding of relative normative relationships and calibration reflecting its ability to map these relationships onto the appropriate numerical scale. Systematic training differences between models may also contribute to these performance patterns. However, examination of these hypotheses would require more detailed information about training procedures than is available to us.

**Comparison of Divergent Scenarios**

To understand if different AI systems make similar kinds of errors, we analyzed the consistency of their most divergent scenarios. In Table 3, we used the top 10 list of errors from GPT-4.5 (see Table 1) as a reference set and examined where these specific scenarios ranked in terms of error magnitude for the next-generation models.

**Table 3. All Models' Prediction Errors for GPT-4.5's Top 10 Divergent Scenarios.**

| Scenario | GPT-4.5 | GPT-5 | Claude Sonnet 4 | Gemini 2.5 Pro |
|---|---|---|---|---|
| Wave to a friend in one's own room | -4.70 | 1.13 | 0.13 | 0.70 |
| Read in church | -3.70 | 1.62 | -4.08 | -1.51 |
| Run at a football game | 3.22 | 1.14 | 3.30 | 1.97 |
| Read on a downtown sidewalk | 3.19 | 2.22 | -1.20 | 3.07 |
| Run in one's own room | 2.88 | 1.99 | 2.70 | 1.07 |
| Kiss at the movies | 2.84 | 2.96 | 2.18 | 1.54 |
| Mumble at the movies | -2.73 | -2.08 | -2.75 | -2.96 |
| Read in class | -2.60 | -0.19 | -5.74 | -4.86 |
| Drink water at a job interview | 2.46 | 2.89 | 1.46 | 1.96 |
| Write at a job interview | 2.44 | 2.51 | -3.69 | 3.06 |

Note: The same scenarios as in Table 1. Entries are differences between the model prediction and the human consensus on the 0-9 scale. Negative values means that the model underestimated the appropriateness of the scenario.

Table 3 indicates a notable overlap in the scenarios that proved most challenging. The prediction errors of the new models were mostly in the same direction as the original GPT-4.5 error. This indicates that while overall accuracy is high, the boundaries of AI's social understanding are consistent, likely reflecting a shared limitation in reasoning about specific, experiential social contexts. To test the systematic nature of these errors more rigorously, we also calculated the correlations between each model's signed error terms (i.e., the deviation from the human average). As shown in Table 4, the error patterns were highly correlated across all models, indicating that when one model over- or underestimated the appropriateness of a given scenario, the other models tended to make an error in the same direction. Thus, while their overall accuracy varies, the leading AI models share a consistent set of limitations in their understanding of human social norms. We will return to what these limitations might be in the General Discussion.

**Table 4. Correlation Matrix of AI Models' Signed Error Terms.**

|  | GPT-4.5 | GPT-5 | Claude Sonnet 4 | Gemini 2.5 Pro |
|---|---|---|---|---|
| GPT-4.5 | — | | | |
| GPT-5 | 0.48 | — | | |
| Claude Sonnet 4 | 0.54 | 0.29 | — | |
| Gemini 2.5 Pro | 0.60 | 0.58 | 0.46 | — |

Note: Each cell contains the Pearson correlation coefficient (r) between the signed error terms of two models across the 555 scenarios.

## General Discussion

This two-part study provides the first quantitative demonstration that artificial intelligence, through statistical learning alone, can achieve a level of performance at the task of predicting human social consensus that exceeds nearly all individual humans on everyday norms that are typically unwritten rules. Across two studies, we found that multiple AI models can predict the consensus of human social norms with an accuracy exceeding the vast majority of individual humans. This result does not necessarily imply a human-like social intelligence per se, but it demonstrates a striking proficiency in a specific meta-cognitive task: to estimate the central tendency of collective human judgment of everyday behavior in various contexts. The fact that this capability can be achieved through statistical learning alone is the primary focus of our discussion.

Study 1 established a powerful proof-of-concept, showing that a state-of-the-art AI (GPT-4.5) could predict U.S. social norms with remarkable accuracy ($R^2 = 0.89$). Study 2 then demonstrated the robustness of this finding across multiple architectures, revealing a complex performance landscape. While a next-generation model (GPT-5) achieved the highest correlation with human consensus ($R^2 = 0.91$), its predecessor (GPT-4.5) demonstrated a lower average error, suggesting that different performance metrics capture different facets of AI social intelligence. The results also show that the limitations of their social understanding are systematic and consistent. Together, these findings have important implications for theories of social cognition and computational approaches to understanding human social behavior.

### Implications for Social Learning Theory

The consistent high performance of multiple AI models, with several achieving correlations near or above $R^2 = 0.90$, provides strong support for bottom-up theories of social learning. While many theories emphasize explicit instruction and social feedback (Bandura, 1977), our findings align more closely with statistical learning approaches suggesting that complex social knowledge can be acquired through pattern extraction from environmental input (Saffran et al., 1996; Elman, 1990). The AIs' performance is consistent with distributional theories proposing that cultural knowledge emerges from co-occurrence patterns in language (Harris, 1954; Landauer & Dumais, 1997) and Bayesian models of pragmatic understanding through probabilistic inference over communicative patterns (Goodman & Frank, 2016).

The successful replication across different architectures strengthens the claim that language contains rich, structured information about social appropriateness. Indeed, the fact that an understanding of norms exceeding the predictive accuracy of individual humans can be achieved from linguistic data alone suggests that language may serve as a more systematic and robust repository for cultural transmission than many theories of cultural evolution typically assume. The AI's superior ability to predict consensus, despite lacking the varied individual experiences that shape human judgment, implies that collective cultural knowledge may be more structured and accessible than the noise of individual variation might suggest.

**Computational Models of Social Cognition**

From a computational perspective, our results reveal insights about the requirements for social understanding. The qualitative differences between models are particularly valuable for understanding how different approaches shape social cognition. For instance, Claude Sonnet 4's "quantized" response pattern suggests a more discrete internal representation of appropriateness, while the GPT models' continuous estimates indicate more analog processing of social gradations. This demonstrates that multiple computational approaches can achieve high-level social understanding, even if their internal representations vary substantially.

A key insight into the nature of AI social cognition comes from the divergence in performance between correlation ($R^2$) and mean absolute error (MAE). The results show that GPT-5 achieved the highest correlation, suggesting it has the most refined model of the *relative relationships* between social norms—it best understands the 'shape' of the normative landscape. However, its predecessor, GPT-4.5, achieved a lower MAE, indicating it is better *calibrated*, with its predictions being, on average, closer to the precise human consensus. This finding is significant from a cognitive science perspective, as it suggests that the ability to understand normative structures (measured by high $R^2$) and the ability to provide perfectly calibrated estimates (measured by low MAE) are distinct and dissociable computational capabilities. That these abilities may not improve in lockstep during model development challenges a monolithic view of "social understanding" and implies that different computational processes may underlie these different facets of social cognition.

The individual-level benchmarking methodology we introduce offers a powerful new tool for evaluating computational models of social cognition. By comparing AI performance to the full distribution of human variation, we can assess whether an artificial system's performance falls within the typical range of human variation or exceeds it—a distinction crucial for understanding the computational basis of social intelligence and for developing more sophisticated theories of cultural competence.

**Understanding the Boundaries of Statistical Social Learning**

The error patterns were consistently positively correlated across all models, indicating that when one model made an error on a scenario, the other models tended to err in the same direction. A plausible interpretation is that certain aspects of social understanding may require computational mechanisms beyond pattern recognition over text. Our analysis of divergent scenarios provides suggestive evidence about the boundaries of statistical social learning, though these interpretations remain necessarily speculative without direct access to model internals.

A clear illustration emerges from the scenario of reading in church, where GPT-4.5, Claude Sonnet 4, and Gemini 2.5 Pro all significantly underestimated its appropriateness (human mean = 6.53, AI range = 2.35 - 4.92). This convergent error plausibly reflects a challenge in resolving semantic ambiguity that depends heavily on contextual interpretation. The models appear to activate a general script for "reading" as a solitary, potentially disruptive activity during a religious service. In contrast, human participants likely interpreted the phrase through a more context-specific lens, recognizing "reading in church" as referring to the communal and appropriate reading of religious texts such as hymns, prayers, or scripture. This suggests that complete social understanding may require flexible contextual reasoning that goes beyond pattern recognition over linguistic co-occurrences.

Other systematic errors point to different types of cognitive challenges. All models overestimated the appropriateness of kissing at the movies (human mean = 4.57, AI range = 6.11 - 7.53), which may reflect training data biases where literary and media representations over-emphasize romantic behaviors in cinema settings while underrepresenting their real-world social costs to bystanders.

Conversely, the models consistently underestimated the appropriateness of mumbling at the movies (human mean = 4.00, AI range = 1.04 - 1.92). While "mumbling" typically carries negative connotations, in the specific context of a movie theater it represents the most socially appropriate way to communicate without disturbing others. The models appear to apply the behavior's general negative valence without recognizing this crucial context-dependent exception.

These examples illustrate three potential boundaries of pattern-based social understanding. First, semantic ambiguity resolution may require experiential knowledge about how behaviors are actually performed in specific contexts rather than just their linguistic descriptions. Second, training data biases may systematically misrepresent certain social situations, particularly when media portrayals diverge from real-world social norms. Third, context-dependent valence shifts may be difficult to capture through statistical patterns alone, as they require understanding how situational factors can override general behavioral associations.

What is particularly remarkable is that these systematic limitations do not substantially compromise the models' overall performance. Despite these specific challenges, all models achieved correlations above $R^2 = 0.82$ and outperformed the vast majority of individual humans, suggesting that statistical learning can yield robust social understanding even with identifiable gaps. However, the persistence of these limitations across different architectures and the evolution to more sophisticated models suggests they may represent genuine challenges for statistical approaches to social cognition.

**Limitations and Future Directions**

An important limitation of this study is that the norms were restricted to the U.S. cultural context. While the success of multiple models in this context provides a strong foundation for understanding statistical approaches to cultural learning, systematic cross-cultural investigations are needed to test the generalizability of these computational mechanisms across different cultural contexts. Cross-cultural research demonstrates variation in everyday norms (Gelfand et al., 2011), and it is possible that these models would not be equally successful in predicting culturally distant societies' everyday norms. This is an area for future research.

Another limitation is that our individual-level benchmarking compared an AI performing an explicit prediction task with humans performing a subjective judgment task. While we argue these tasks are comparable for assessing an understanding of collective norms, future research could directly compare AI and human performance when both are given the identical meta-cognitive task of predicting group averages.

## Conclusion

Our two-part study demonstrates that sophisticated models that can predict human social norm consensus can emerge from statistical learning over linguistic data alone, with multiple AI architectures achieving an accuracy in predicting cultural consensus that exceeds the vast majority of individual humans. While the models performed a meta-cognitive prediction task distinct from the humans' subjective rating task, this ability to model the cultural consensus from linguistic data alone has profound implications for theories of social learning. These findings support bottom-up theories of social learning while revealing systematic boundaries of pattern-based social understanding. The consistent limitations across different computational approaches suggest that complete social cognition may require integration of statistical, experiential, and embodied forms of knowledge.

From a cognitive science perspective, these results suggest that cultural knowledge representation may be more amenable to computational approaches than previously thought, while simultaneously highlighting the computational challenges involved in achieving fully human-like social understanding. The methodological innovation of individual-level benchmarking provides a new tool for investigating the computational basis of social cognition and for developing more sophisticated theories of how cultural competence emerges and operates.

## Acknowledgments

Large language models (Google's Gemini 2.5 Pro and Anthropic's Claude Sonnet 4) provided helpful suggestions for improving the flow of the text.

## Data Availability Statement

All data and analysis code are available at OSF
https://osf.io/mydqf/?view_only=ac212881f247416c98e1baac06a478ac

# Supplementary Material

**Supplementary Figure 1**. Boxplots showing the difference between individual human MAE and LLM MAE when calculated on identical scenario subsets. For each human participant, we computed MAE using only the scenarios they rated and compared this to the LLM's MAE on the same scenarios. Positive values indicate the LLM outperformed the human participant. The predominantly positive differences confirm that LLMs outperformed nearly all individual humans.

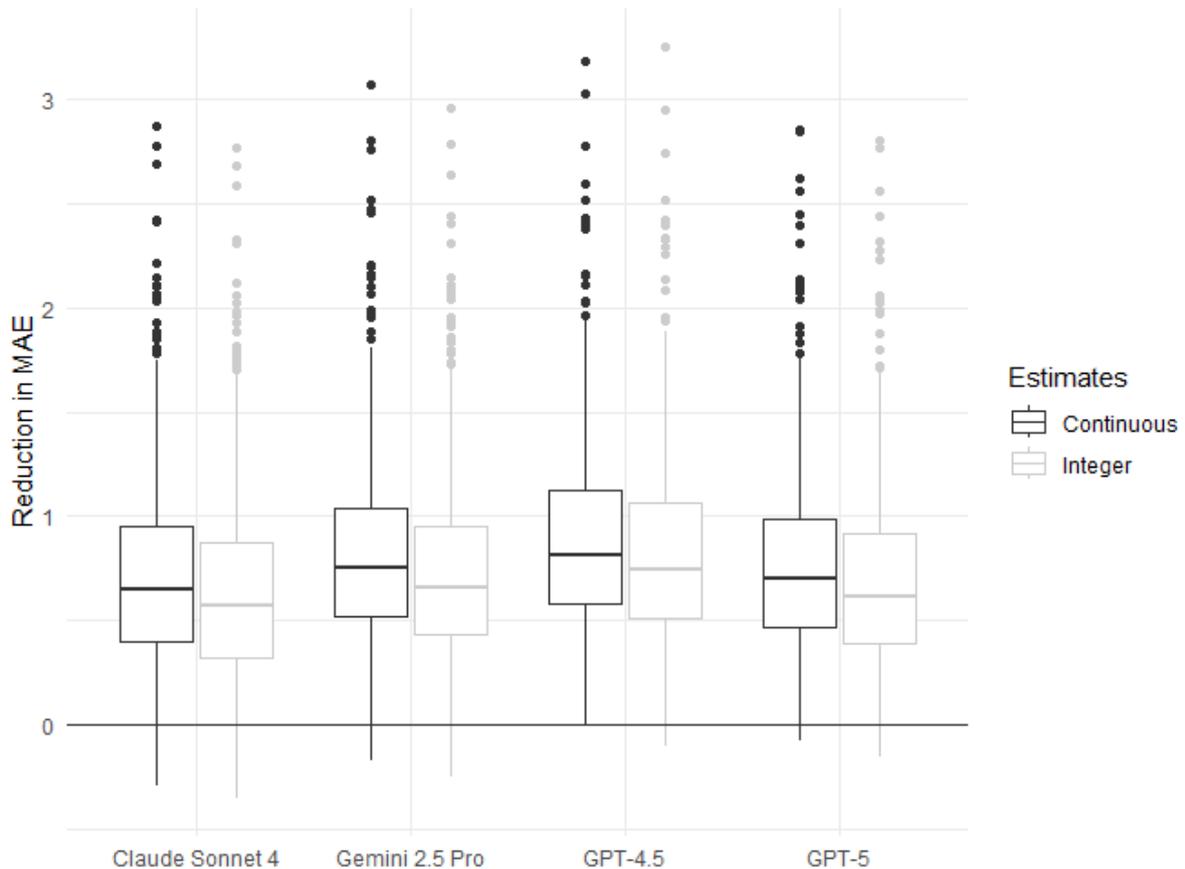